\documentclass[10pt,twocolumn,letterpaper]{article}
\usepackage[pagenumbers]{cvpr}
\usepackage{amsmath,amssymb,booktabs,multirow,array,xspace}
\usepackage{graphicx,xcolor}
\usepackage{microtype}
\definecolor{draftblue}{RGB}{30,85,150}
\newcommand{\method}{Diffutoon-2\xspace}

\usepackage{longtable}

\usepackage{algorithm}
\usepackage{algpseudocode}
\algrenewcommand\alglinenumber[1]{\scriptsize #1}
\algrenewcommand\algorithmicrequire{\textbf{Input}}
\algrenewcommand\algorithmicensure{\textbf{Output}}

\usepackage{tabularx}
\newcolumntype{Y}{>{\centering\arraybackslash}X}
\usepackage{listings}
\lstdefinestyle{captionprompt}{
  basicstyle=\ttfamily\scriptsize,
  columns=fullflexible,
  keepspaces=true,
  breaklines=true,
  breakatwhitespace=false,
  showstringspaces=false,
  frame=single,
  framerule=0.3pt,
  aboveskip=0.6em,
  belowskip=0.6em
}

\usepackage[most]{tcolorbox}
\DeclareCaptionType{prompt}[Prompt][List of Prompts]
\tcbset{promptbox/.style={enhanced,colback=black!2!white,
 colframe=black!35,boxrule=0.5pt,arc=1mm,
 left=3mm,right=3mm,top=2mm,bottom=2mm,before skip=3pt,after skip=0pt}}
\newcommand{\promptfile}[3]{%
 \par\addvspace{12pt}\noindent\begin{minipage}{\textwidth}
 \captionof{prompt}{#2}\label{#1}
 \tcbinputlisting{promptbox,listing only,listing file={#3},
 listing options={basicstyle=\ttfamily\small,columns=fullflexible,
 keepspaces=true,breaklines=true,breakatwhitespace=false,
 showstringspaces=false,literate={—}{{\textemdash}}1 {–}{{\textendash}}1}}
 \end{minipage}\par
}

\definecolor{captiongreen}{RGB}{22,112,55}
\definecolor{captionred}{RGB}{176,40,40}
\newcommand{\captionmatch}[1]{\textcolor{captiongreen}{\textbf{#1}}}
\newcommand{\captiondrift}[1]{\textcolor{captionred}{\textbf{#1}}}

\definecolor{cvprblue}{rgb}{0.21,0.49,0.74}
\usepackage[pagebackref,breaklinks,colorlinks,allcolors=cvprblue]{hyperref}
\title{Diffutoon-2: Toon Shading with Video Diffusion Models}
\author{
    Yuze Sun\textsuperscript{1,2},
    Zhongjie Duan\textsuperscript{2},
    Yingda Chen\textsuperscript{2} and
    Jingjing Chen\textsuperscript{3}\\
    \textsuperscript{1}Shanghai Jiao Tong University \\
    \textsuperscript{2}ModelScope Team, Alibaba Group \\
    \textsuperscript{3}Fudan University\\ 
}
\makeatletter
\apptocmd{\@maketitle}{
\noindent\begin{minipage}{\textwidth}
\centering
\captionsetup{hypcap=false}
\includegraphics[width=\textwidth]{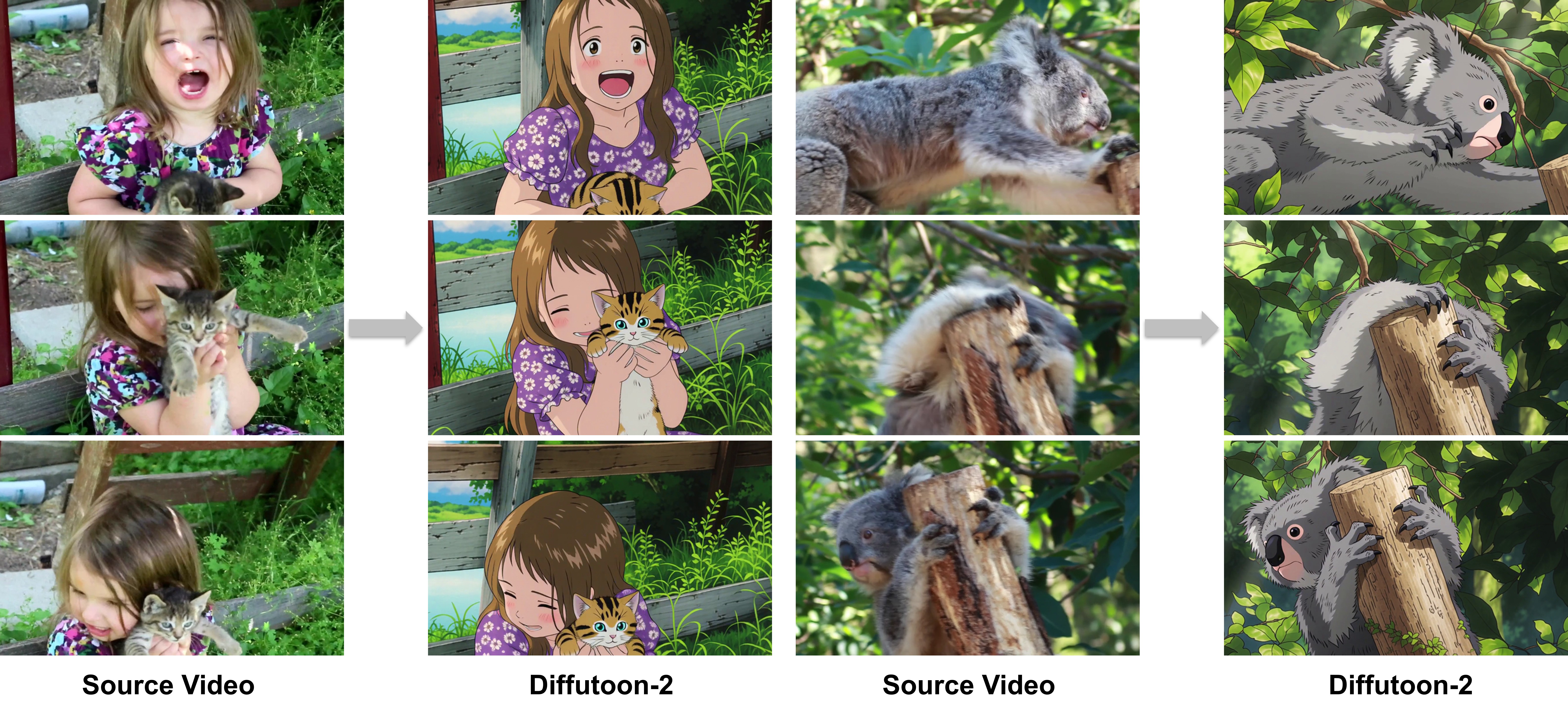}
\captionof{figure}{Toon shading results of \method. Each example pairs source frames with their animated counterparts at corresponding timestamps. The target appearance combines clean contours and simplified shading with the composition and color identity of the source.}
\label{fig:teaser}
\end{minipage}
\par\vspace{12pt}
}{}{\errmessage{Could not insert teaser}}
\makeatother
\begin{document}
\maketitle
\begin{abstract}
Toon shading renders objects with a flat, stylized appearance. Diffusion models open a new route to this task by transforming photorealistic videos into an anime style. However, existing approaches often rely on curated source and style training pairs or additional visual references. Diffutoon avoids these requirements through structure guidance and colorization, but builds on an image diffusion backbone with short-window motion modules and repeated overlapping-window denoising. We present Diffutoon-2, a video diffusion framework that learns toon shading through animation reconstruction. Two complementary modules independently reconstruct animation clips from outlines and blurred colors, with the video backbone frozen. At inference, the modules jointly guide generation using the same types of conditions extracted from photorealistic footage, requiring neither cross-domain training pairs nor external style references. Extensive experiments demonstrate that \method outperforms both general-purpose and domain-specific baselines, across quantitative metrics and human evaluation. All resources will be open sourced\footnote{\url{https://github.com/modelscope/DiffSynth-Studio}}, including code and models.
\end{abstract}

\section{Introduction}
\label{sec:intro}

Toon shading is a non-photorealistic rendering task that gives objects a flat, stylized appearance and is widely used in animation production. With the rise of diffusion models, video stylization has provided ways to achieve this appearance from photorealistic footage by rendering it in anime style. The source video supplies the subjects, composition, and motion, while the model transforms texture and shading into the clean outlines and simplified color regions characteristic of anime rendering. This formulation brings toon shading to recorded scenes without requiring explicit 3D representation.

Existing approaches obtain the target appearance through paired supervision or visual references. EchoStyle~\cite{echostyle} constructs source and style video pairs, while PickStyle~\cite{pickstyle} uses paired images to synthesize aligned training clips. The reference-guided formulation of Kiwi-Edit~\cite{kiwiedit} uses reference images during training and inference, whereas AnyV2V~\cite{anyv2v} edits the first source frame and propagates its appearance through video generation. These routes introduce two dependencies. Paired supervision requires curated examples that change style while preserving content correspondence. Reference-based generation requires a suitable image whose quality and semantic coverage affect subsequent rendering. In particular, first-frame editing errors can propagate through a video, and scene changes can introduce content that the initial reference does not represent. Diffutoon~\cite{diffutoon} repurposes off-the-shelf modules designed for image generation from line art and image super-resolution to provide structure guidance and colorization. However, its image diffusion backbone relies on motion modules with windows of at most 32 frames, while repeated denoising of overlapping windows adds computational cost.

We present \method, a framework that renders photorealistic videos in an anime style by conditioning a video diffusion model on outlines and blurred colors extracted from the input video. The outline video constrains scene structure, while the blurred color video preserves spatial color information and suppresses photographic texture. Both types of supervision are readily available from animation footage itself, since line-art extraction and Gaussian blur each produce a condition paired with the original animation. We use these two forms of supervision to train the structure and colorization modules through animation reconstruction, avoiding the need to generate aligned photorealistic counterparts. The modules are optimized independently with the same backbone kept frozen and composed through residual fusion at inference. At inference, both conditions come directly from the source photorealistic video, eliminating external style references and a separate first-frame editing stage. Generated context supports continuation beyond a single generation window.

Our contributions are summarized as follows.
\begin{itemize}
\item We present \method, a technique for toon shading with video diffusion models that renders photorealistic videos in an anime style through structure guidance and colorization.
\item We formulate complementary animation reconstruction tasks for structure and colorization, enabling independent training and joint control at inference without cross-domain paired supervision or external reference images.
\item On extensive experiments, \method outperforms both general-purpose and domain-specific baselines, including the closed-source Seedance2.5, across quantitative metrics and human evaluation.
\end{itemize}
\section{Related Work}
\paragraph{Cartoonization and animation colorization.}
Generative animation covers stylization, colorization, and motion~\cite{celsurvey}. Cartoonization models contours and simplified appearance~\cite{cartoongan,whitebox}; diffusion methods colorize animation line art from references~\cite{animediffusion,animecolor}, propagate colors through grayscale videos~\cite{colordiffuser}, and render videos~\cite{diffsynth,diffutoon}. FlowStyler~\cite{flowstyler} transports geometric and color fields. We learn outline and low-frequency color guidance through animation reconstruction.

\paragraph{Video stylization and editing.}
Temporal propagation and synchronization support diffusion editing~\cite{rerender,tokenflow,anyv2v,flowvid,smfd,rfsolver}, including local guidance for faces~\cite{lcg}. Stylization uses text~\cite{styleavideo,vstylist}, visual references~\cite{stylecrafter,stylemaster,telestyle}, or broader conditions and modalities~\cite{dreamstyle,unistylediff}. Paired supervision supports both stylization~\cite{pickstyle,echostyle} and editing~\cite{ditto,easyv2v,kiwiedit}. Meet-in-Style and StructuReiser~\cite{meetinstyle,structureiser} learn fast networks from stylized examples, with StructuReiser emphasizing structure preservation. We render directly with video diffusion trained through animation-only reconstruction.

\paragraph{Controllable video generation.}
Video diffusion supports diverse controls~\cite{videosurvey}, including spatial residuals~\cite{controlnet}, motion and appearance for image animation~\cite{animatediff,animateanyone,dynamicrafter}, and trajectories for anime synthesis~\cite{trajectoryanime}. DiTCtrl~\cite{ditctrl} controls attention across prompts; V-RGBX~\cite{vrgbx} edits intrinsic scene properties. RTR-DiT~\cite{rtrdit} enables streaming stylization with autoregressive distillation and reference-preserving caching. We combine outline and low-frequency color guidance with clean latent context for window continuation.

\section{Method}
\label{sec:method}

\begin{figure*}[t]
\centering
\includegraphics[width=\textwidth]{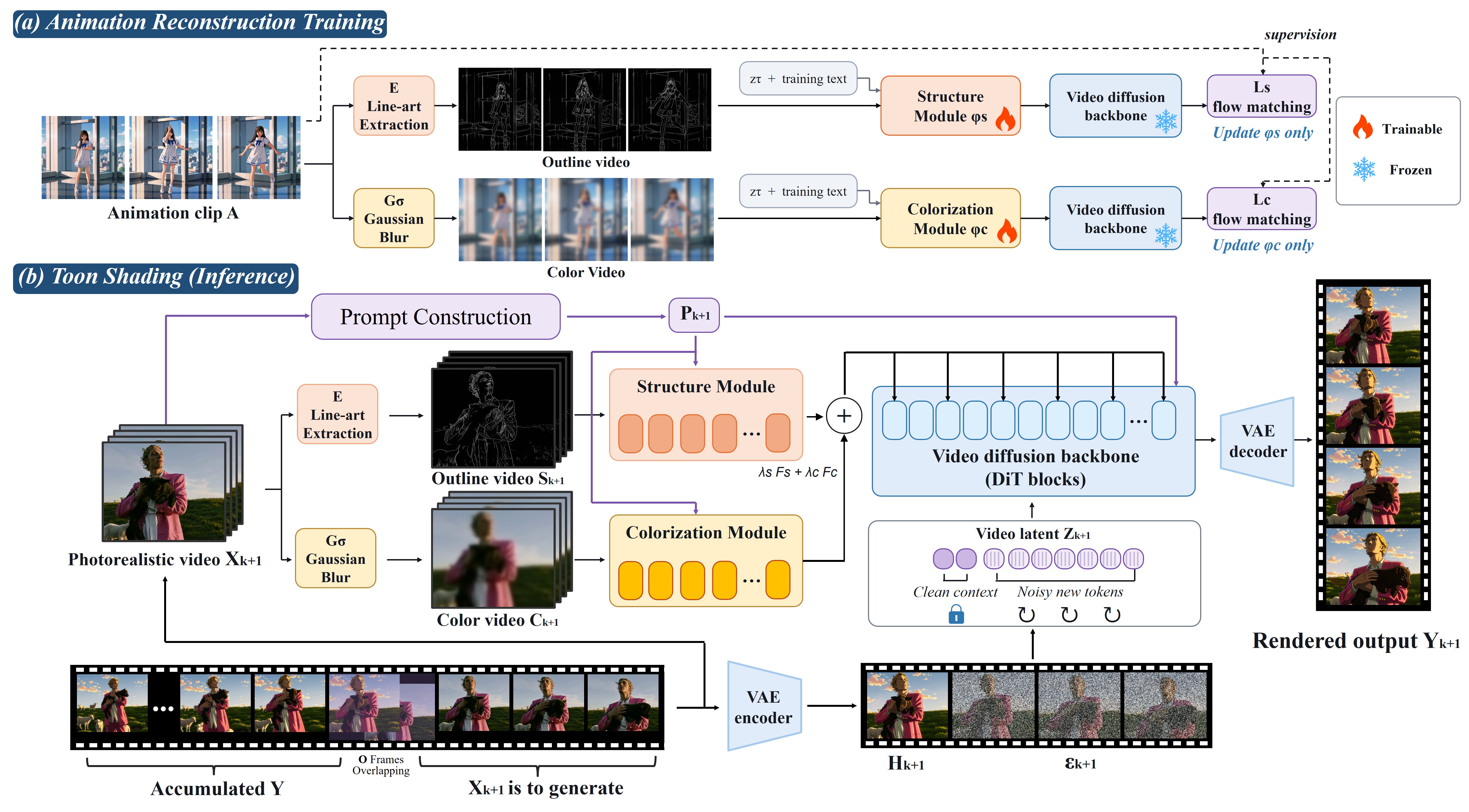}
\caption{Overview of \method. (a) The structure and colorization modules are trained independently to reconstruct animation from outlines and blurred colors, with the video backbone frozen. (b) During inference, both modules guide video generation using source-derived conditions and constructed prompts. Clean context latents are retained across windows while new-region latents are denoised.}
\label{fig:pipeline}
\end{figure*}

Figure~\ref{fig:pipeline} provides an overview of our framework. We first formalize video toon shading as conditional video generation that renders photorealistic inputs in an anime style while preserving their content. We then describe the structure and colorization modules, which learn animation reconstruction from complementary conditions and jointly guide the video diffusion backbone. Finally, we introduce window continuation and prompt construction for rendering videos beyond a single generation window.

\subsection{Problem Formulation}
\label{sec:formulation}
Given a photorealistic video with $T$ frames $X=\{x_i\}_{i=1}^{T}$, we aim to generate an output $Y=\{y_i\}_{i=1}^{T}$ in an anime style while preserving the source composition, colors, and motion. Let $E$ denote a line-art extractor and $G_\sigma$ a Gaussian blur. The two visual conditions are
\begin{equation}
 S=E(X),\qquad C=G_\sigma(X),
\end{equation}
where $S$ is the outline video, $C$ is the color video, and $\sigma$ is the standard deviation of the Gaussian blur. A textual condition $p$ specifies the target style and the content of the input.

For a video partitioned into $K$ generation windows indexed by $k=0,\ldots,K-1$, let $X_k$, $S_k$, and $C_k$ denote corresponding source and condition segments, $p_k$ the window-specific text, and $H_k$ the generated context from the preceding window. The first window has $H_0=\varnothing$. We denote the video generator by $\mathcal{D}_{\theta,\phi_s,\phi_c}$, where $\theta$ contains the backbone parameters and $\phi_s,\phi_c$ belong to the structure and colorization modules. With Gaussian latent noise $\epsilon_k$, generation and assembly are written as
\begin{equation}
\begin{aligned}
 Y_k &= \mathcal{D}_{\theta,\phi_s,\phi_c}
       (\epsilon_k,S_k,C_k,p_k,H_k),\\
 Y &= \mathcal{A}_T(Y_0,\ldots,Y_{K-1}),
\end{aligned}
\label{eq:generation}
\end{equation}
where $Y_k$ is the generated output for window $k$ and $T$ is the source frame count. The assembly operator $\mathcal{A}_T$ merges each pair of overlapping frames into one blended frame, appends the non-overlapping frames, and retains the first $T$ output frames. For a single window, this reduces to generation conditioned on the source outlines, colors, and text.

\begin{table*}[t]
\centering
\caption{Quantitative comparison between Diffutoon-2 and other methods. $\downarrow$: lower is better. $\uparrow$: higher is better. The \textbf{best} and \underline{second-best} results are highlighted in \textbf{bold} and \underline{underline}, respectively.}
\label{tab:main}
\scriptsize\setlength{\tabcolsep}{1.5pt}
\begin{tabularx}{\linewidth}{@{}l*{9}{Y}@{}}
\toprule
\multirow{2}{*}{Method}
& \multicolumn{3}{c}{\scriptsize Content Preservation}
& \multicolumn{3}{c}{\scriptsize Style and Text Alignment}
& \multicolumn{3}{c}{\scriptsize Visual and Temporal Coherence}\\
\cmidrule(lr){2-4}\cmidrule(lr){5-7}\cmidrule(lr){8-10}
& \mbox{Line-F1 $\uparrow$} & \mbox{Chroma $\downarrow$} & \mbox{UR-E $\uparrow$}
& \mbox{CLIP $\uparrow$} & \mbox{UR2-S $\uparrow$} & \mbox{UR2-A $\uparrow$}
& \mbox{CSD $\uparrow$} & \mbox{Smooth. $\uparrow$} & \mbox{UR2-C $\uparrow$}\\
\midrule
Diffutoon & \underline{0.60} & 9.30 & \underline{22.00} & 24.60 & 3.22 & 2.93 & 0.80 & \underline{0.98} & 3.64\\
EchoStyle & 0.51 & 7.61 & 21.90 & 24.20 & 3.17 & 3.11 & 0.86 & 0.97 & 3.72\\
PickStyle & 0.35 & 8.98 & 21.80 & 25.80 & \underline{3.37} & 3.10 & 0.87 & 0.97 & \underline{3.82}\\
Kiwi-Edit & 0.52 & 10.97 & 21.80 & 22.90 & 3.24 & 3.05 & 0.86 & \underline{0.98} & 3.66\\
AnyV2V & 0.58 & 8.86 & 15.80 & 13.30 & 3.23 & 2.69 & 0.83 & 0.95 & 3.71\\
MiniMax-H3-Ref2VA & 0.48 & \underline{5.40} & 14.00 & 12.80 & 3.33 & 2.81 & 0.85 & 0.97 & 3.78\\
Seedance2.5 & 0.47 & 6.84 & 21.90 & \underline{26.10} & 3.26 & \underline{3.19} & \underline{0.88} & \textbf{0.99} & 3.75\\
\midrule
\method & \textbf{0.64} & \textbf{4.86} & \textbf{23.00} & \textbf{30.50} & \textbf{3.38} & \textbf{3.39} & \textbf{0.92} & \textbf{0.99} & \textbf{3.87}\\
\bottomrule
\end{tabularx}
\end{table*}

\subsection{Structure Module}
\label{sec:structure}
We use the line-drawing extractor of Informative Drawings~\cite{informative_drawings} to obtain $S=E(X)$ frame by frame. The extractor produces contours that convey scene geometry and semantics while suppressing surface texture. The extracted drawings are resized to the source frame dimensions.

For training, an animation clip $A$ provides the outline condition $E(A)$ and the reconstruction target $A$, together with the structure-training text $p_s$. We optimize the structure module with a flow-matching loss~\cite{flowmatching},
\begin{equation}
 \mathcal{L}_s=\mathbb{E}_{A,\epsilon,\tau}
 \left[\left\|v_{\theta,\phi_s}(z_\tau,\tau,E(A),p_s)-u\right\|_2^2\right],
\label{eq:structure_loss}
\end{equation}
where $v_{\theta,\phi_s}$ is the velocity predicted by the frozen backbone conditioned by the structure module. The noisy latent is $z_\tau=(1-\tau)z_A+\tau\epsilon$, with $z_A$ the encoded animation, $\epsilon\sim\mathcal{N}(0,I)$ Gaussian noise, and $\tau\in[0,1]$ sampled according to the backbone scheduler. Here $\tau=0$ corresponds to the clean animation latent and $\tau=1$ to pure Gaussian noise. The target velocity is $u=\epsilon-z_A$, and only $\phi_s$ is optimized.

At inference, the trained module receives the source outlines and outputs $\mathcal{F}_{\phi_s}^{\ell}(z_\tau,\tau,S,p)$, the structural residual feature at backbone layer $\ell$. Here $z_\tau$ is the current generation latent and $p$ is the inference text. These features constrain the spatial structure of the rendered video.

\subsection{Colorization Module}
\label{sec:colorization}
We construct the color video by applying spatial Gaussian filtering independently to each input frame,
\begin{equation}
\begin{aligned}
 C^{(i)}(\mathbf{q}) &= (K_\sigma*x_i)(\mathbf{q}),\\
 K_\sigma(\boldsymbol{\delta})
 &=\frac{1}{2\pi\sigma^2}
   \exp\!\left(-\frac{\|\boldsymbol{\delta}\|_2^2}{2\sigma^2}\right),
\end{aligned}
\label{eq:gaussian}
\end{equation}
where $C^{(i)}$ is the filtered frame corresponding to $x_i$, $\mathbf{q}$ is a spatial position, $\boldsymbol{\delta}$ is a spatial offset, $K_\sigma$ is the normalized Gaussian kernel, and $*$ denotes spatial convolution. The resulting condition preserves broad color regions while attenuating photographic texture, without filtering across time.

For training, we use the pair $(G_\sigma(A),A)$ with colorization text $p_c$ and optimize the flow-matching loss
\begin{equation}
 \mathcal{L}_c=\mathbb{E}_{A,\epsilon,\tau}
 \left[\left\|v_{\theta,\phi_c}(z_\tau,\tau,G_\sigma(A),p_c)-u\right\|_2^2\right],
\label{eq:color_loss}
\end{equation}
where $v_{\theta,\phi_c}$ is the velocity predicted with the colorization module, using the same latent path and target velocity as in Eq.~\ref{eq:structure_loss}. Only $\phi_c$ is optimized in this run. The two modules are trained independently on the same animation dataset, with the backbone and encoders frozen. Each training run includes only one control module, avoiding the simultaneous storage of gradients and optimizer states for both modules. At inference, the colorization module receives the blurred input video $C$ and outputs $\mathcal{F}_{\phi_c}^{\ell}(z_\tau,\tau,C,p)$, the color residual feature at layer $\ell$.

\paragraph{Joint inference.}
The structure and colorization modules jointly condition the video backbone on the outline video $S$ and color video $C$. Both modules use the shared text $p$, and their residual features are combined at each controlled layer as
\begin{equation}
 r_\ell=\lambda_s\mathcal{F}_{\phi_s}^{\ell}(z_\tau,\tau,S,p)
       +\lambda_c\mathcal{F}_{\phi_c}^{\ell}(z_\tau,\tau,C,p),
\label{eq:control}
\end{equation}
where $r_\ell$ is the combined residual injected into the backbone and $\lambda_s,\lambda_c$ are the control strengths. For windowed rendering, the conditions and text are replaced by $S_k$, $C_k$, and $p_k$.

\subsection{Window Continuation}
\label{sec:consistency}
We process the video across $K$ windows of at most $W$ frames with overlap $O$ and stride $W-O$. The first window ($k=0$) is generated from noise. For each subsequent window ($k>0$), we take the final $O$ frames of the accumulated output $Y$ as context $H_k$ and retain their clean latent representation. Only the remaining region, corresponding to $W-O$ new frames for a full window, is generated from noise.

At every denoising step, the model receives clean context latents followed by the current latents of the new region,
\begin{equation}
\begin{aligned}
 \widetilde z &= \operatorname{Concat}(z_{\mathrm{ctx}},z^{\mathrm{new}}),\\
 t_j &=
 \begin{cases}
 1, & j\in\mathcal{I}_{\mathrm{ctx}},\\
 1-\tau, & j\in\mathcal{I}_{\mathrm{new}}.
 \end{cases}
\end{aligned}
\label{eq:anchor}
\end{equation}
Here $\widetilde z$ is the full-window latent input to the model, $z_{\mathrm{ctx}}$ denotes the clean encoded context, $z^{\mathrm{new}}$ denotes the evolving new-region latent, and $\operatorname{Concat}$ concatenates them along the temporal dimension. The sets $\mathcal{I}_{\mathrm{ctx}}$ and $\mathcal{I}_{\mathrm{new}}$ index their video tokens after temporal VAE compression. The model timestep $t_j$ uses the reverse convention, with $t_j=0$ denoting pure noise and $t_j=1$ the clean latent. Context tokens use $t_j=1$, while new tokens use $t_j=1-\tau$. Only tokens in $\mathcal{I}_{\mathrm{new}}$ are updated.

Both modules receive the full current-window conditions $S_k$ and $C_k$, including the overlap and new frames. Algorithm~\ref{alg:continuation} summarizes the inference procedure. The descending noise schedule is $\tau_0,\ldots,\tau_N$, with $N$ denoising steps. The vector $\mathbf{t}$ contains the per-token model timesteps, and $v[\mathcal{I}_{\mathrm{new}}]$ selects the predicted velocities for the new region.

After denoising, we decode the full window latent into RGB frames $Y_k$. For the first window ($k=0$), we initialize $Y\gets Y_0$. For $k>0$, the first $O$ frames of $Y_k$ overlap with the context $H_k$. We merge each pair of overlapping RGB frames via linear blending,
\begin{equation}
\begin{aligned}
 \overline Y_{k,i} &= (1-\alpha_i)H_{k,i}+\alpha_i Y_{k,i},\\
 \alpha_i &=
 \begin{cases}
 i/(O-1), & O>1,\\
 1/2, & O=1.
 \end{cases}
\end{aligned}
\label{eq:overlap_blend}
\end{equation}

Here $H_{k,i}$ and $Y_{k,i}$ denote frame $i$ of $H_k$ and $Y_k$, respectively, and $\overline Y_{k,i}$ is the blended frame, with $i=0,\ldots,O-1$. The coefficient $\alpha_i$ is the blending weight of the current-window frame. We then update $Y$ by replacing its last $O$ frames with $\{\overline Y_{k,i}\}_{i=0}^{O-1}$ and appending the remaining new frames of $Y_k$.

\begin{algorithm}[t]
\caption{Continuous Generation at Inference}
\label{alg:continuation}
\small
\begin{algorithmic}[1]
\Require Conditions $\{S_k,C_k,p_k\}_{k=0}^{K-1}$, overlap $O$
\Require Noise schedule $\{\tau_n\}_{n=0}^{N}$
\Ensure Rendered video $Y$ with $T$ frames
\State $Y\gets\varnothing$
\For{$k=0,\ldots,K-1$}
  \If{$k=0$}
    \State $z_{\mathrm{ctx}}\gets\varnothing$
  \Else
    \State $H_k\gets\text{last }O\text{ frames of }Y$
    \State $z_{\mathrm{ctx}}\gets\Call{EncodeContext}{H_k}$
  \EndIf
  \State Identify latent-token index sets $\mathcal{I}_{\mathrm{ctx}}$ and $\mathcal{I}_{\mathrm{new}}$
  \State $z^{\mathrm{new}}\sim\mathcal{N}(0,I)$
  \For{$n=0,\ldots,N-1$}
    \State $\widetilde z\gets\operatorname{Concat}(z_{\mathrm{ctx}},z^{\mathrm{new}})$
    \For{each latent-token index $j\in\mathcal{I}_{\mathrm{ctx}}\cup\mathcal{I}_{\mathrm{new}}$}
      \If{$j\in\mathcal{I}_{\mathrm{ctx}}$}
        \State $t_j\gets1$
      \Else
        \State $t_j\gets1-\tau_n$
      \EndIf
    \EndFor
    \State $v\gets\Call{PredictVelocity}{\widetilde z,\mathbf{t},S_k,C_k,p_k}$
    \State $z^{\mathrm{new}}\gets z^{\mathrm{new}}+(\tau_{n+1}-\tau_n)v[\mathcal{I}_{\mathrm{new}}]$
  \EndFor
  \State $Y_k\gets\Call{Decode}{\operatorname{Concat}(z_{\mathrm{ctx}},z^{\mathrm{new}})}$
  \If{$k=0$}
    \State $Y\gets Y_0$
  \Else
    \State $Y\gets\Call{LinearBlend}{Y,Y_k,O}$
  \EndIf
\EndFor
\State \Return the first $T$ frames of $Y$
\end{algorithmic}
\end{algorithm}

\subsection{Prompt Construction}
\label{sec:prompt}
We use Qwen3.8-27B~\cite{qwen38} to describe video content from sampled frames. For a video that fits within one generation window ($T\leq W$), its caption $d$ is combined with a fixed anime-style instruction $p_{\mathrm{style}}$ to form the generation prompt,
\begin{equation}
 p=\operatorname{Concat}(p_{\mathrm{style}},d),
\label{eq:single_prompt}
\end{equation}
where $\operatorname{Concat}$ denotes text concatenation.

For longer videos ($T>W$), we first describe the whole video and then generate a local caption for each window using the global description as a reference,
\begin{equation}
\begin{aligned}
 d_{\mathrm{global}} &= \mathcal{Q}_{g}(X),\\
 d_{\mathrm{local},k} &= \mathcal{Q}_{l}(X_k,d_{\mathrm{global}}),\\
 p_k &= \operatorname{Concat}(p_{\mathrm{style}},d_{\mathrm{local},k}),
\end{aligned}
\label{eq:captions}
\end{equation}
where $\mathcal{Q}_{g}$ and $\mathcal{Q}_{l}$ are the global and local Qwen captioning procedures. The global caption $d_{\mathrm{global}}$ provides consistent subject names across windows, while $d_{\mathrm{local},k}$ describes the appearance and actions visible in the current segment. Only the local caption is included in $p_k$. Captioning details and examples of improved naming consistency with global guidance are provided in the supplementary material.

\section{Experiments}
\label{sec:experiments}

\begin{figure*}[t]
\centering
\includegraphics[width=\textwidth]{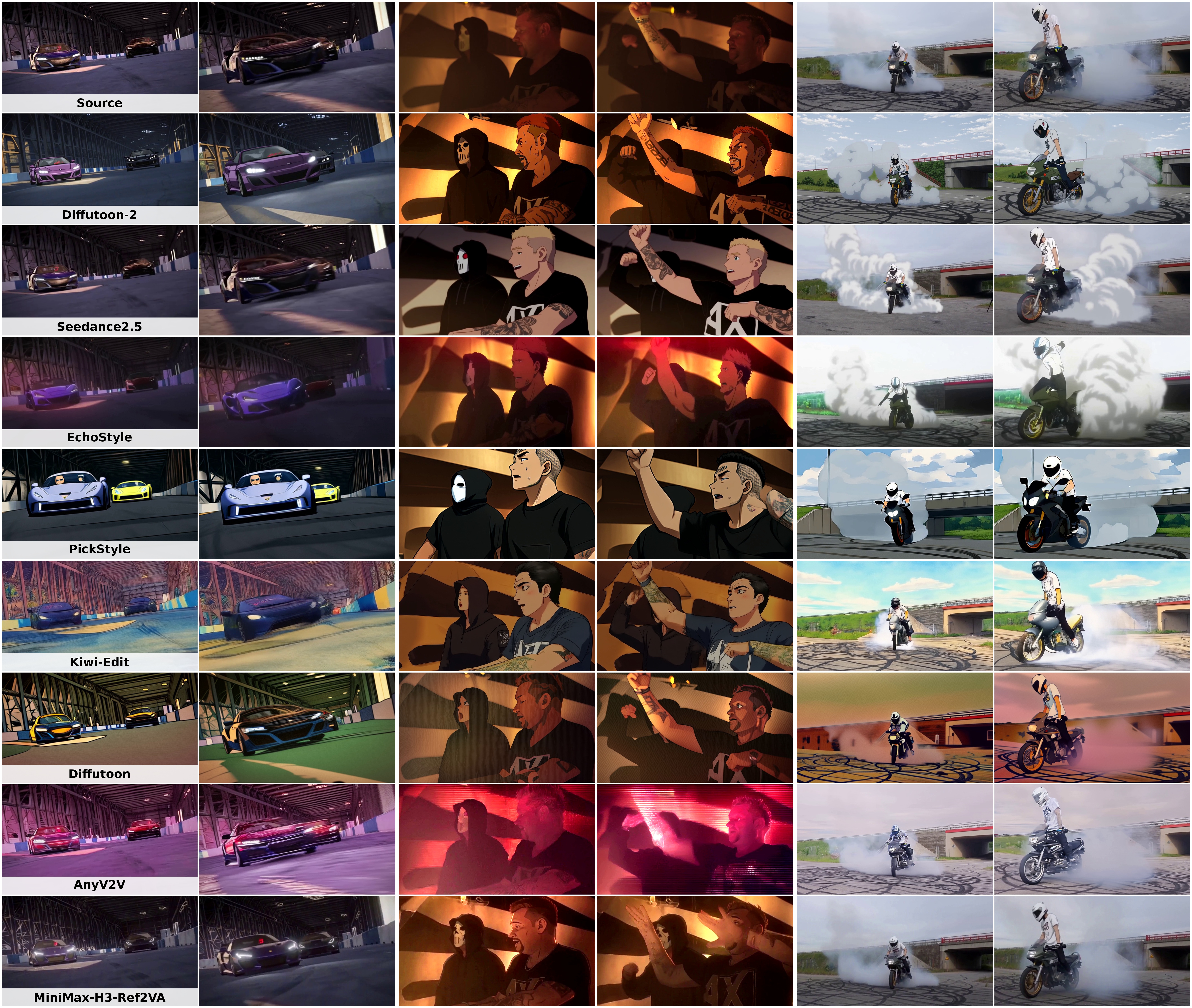}
\caption{Visual comparison of toon shading results. \method produces clear outlines and simplified shading while preserving the source colors and scene structure.}
\label{fig:comparison}
\end{figure*}

\begin{figure*}[t]
\centering
\includegraphics[width=\textwidth]{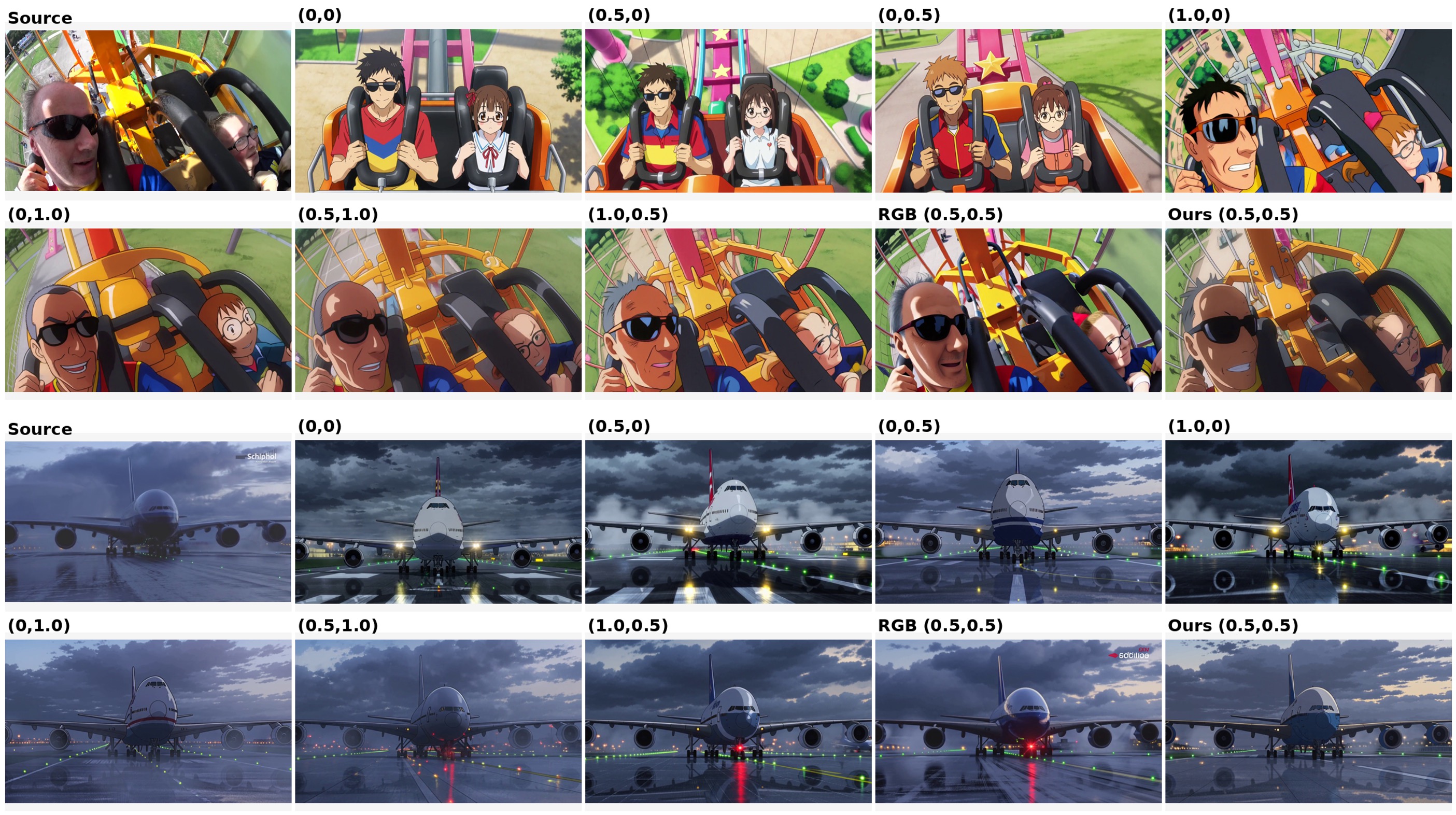}
\caption{Effect of structure guidance, colorization, and the colorization training condition. Matched source intervals and random seeds allow contours, local colors, and appearance to be compared across the variants in Table~\ref{tab:ablation}.}
\label{fig:ablation}
\end{figure*}

\subsection{Implementation Details}
We train separate structure and colorization modules on a private dataset of 15,605 animation clips. We use MiniMax-H3 as the backbone. Both modules adopt a ControlNet architecture~\cite{controlnet}, with nine control blocks corresponding to backbone block IDs $0,5,10,15,20,25,30,35,40$. Each module is trained for 5,000 steps on eight PPU-ZW810s with 96 GB of memory each, taking approximately 36 hours. We use a batch size of 8, a learning rate of $10^{-5}$, and 39-frame clips. Training resolutions preserve the input aspect ratio and are scaled as large as possible without exceeding one million pixels.

At inference, source videos are resampled to 24 fps. For structure guidance, each frame is resized to a short side of 512 pixels before line-art extraction, and the extracted outlines are resized back to the source dimensions. Gaussian filtering of the input video uses $\sigma=20$. Both control strengths are set to 0.5, with a maximum window length of 328 frames and one frame of overlap. We use a LoRA accelerated through distribution matching distillation (DMD) from LightX2V~\cite{lightx2v} for four-step denoising. More details are provided in the supplementary material.

\begin{figure*}[t]
\centering
\includegraphics[width=\textwidth]{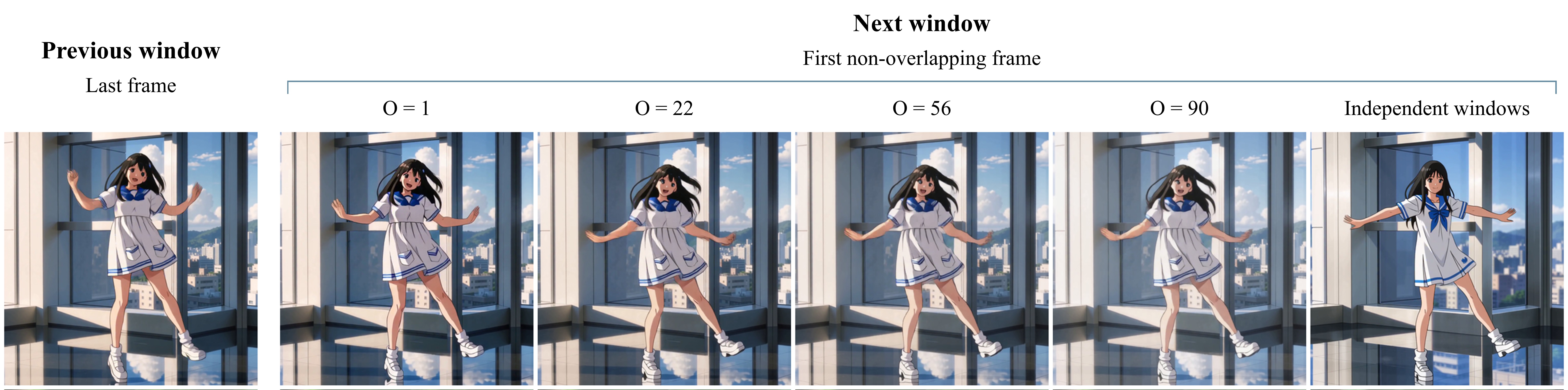}
\caption{Comparison of segment transitions. Each row begins with the last frame of the preceding output window. The remaining columns show the next window's first non-overlapping frame with overlap lengths $O=1,22,56,90$, followed by independent-window generation.}
\label{fig:continuation}
\end{figure*}

\subsection{Evaluation Protocol}
\paragraph{Evaluation data and baselines.}
We evaluate \method on 70 photorealistic videos against both open-source and closed-source systems. We randomly select 30 videos from PixaBay100~\cite{diffsynth} and 40 from the combined training and validation sets of DAVIS2017~\cite{davis2017}. The comparisons include open-source toon shading and video stylization methods (\textbf{Diffutoon}~\cite{diffutoon}, \textbf{EchoStyle}~\cite{echostyle}, and \textbf{PickStyle}~\cite{pickstyle}), open-source video editing and reference-conditioned generation methods (\textbf{Kiwi-Edit}~\cite{kiwiedit}, \textbf{AnyV2V}~\cite{anyv2v}, and \textbf{MiniMax-H3-Ref2VA}), and a general-purpose closed-source model (\textbf{Seedance2.5}). We provide \textbf{MiniMax-H3-Ref2VA} with the same outline and color videos as references, allowing us to assess the benefit of our trained modules over direct reference-conditioned generation. All methods target an anime style and are evaluated on corresponding source intervals.

\paragraph{Evaluation metrics.}
We report nine metrics grouped into content preservation, style and text alignment, and visual and temporal coherence.

\textit{Content preservation.} \textbf{Line-art Boundary F1-score (Line-F1)} measures agreement between outlines extracted from aligned source and generated frames, using binarized line maps and a spatial matching tolerance. \textbf{Chroma} adapts \textbf{CIEDE2000} by removing lightness differences and applying Gaussian smoothing to the chroma channels. This formulation permits changes in luminance associated with cel shading while measuring the spatial agreement of color. \textbf{UnifiedReward-Edit (UR-E)}~\cite{uredit} jointly evaluates editing-instruction adherence and source-content preservation.

\textit{Style and text alignment.} \textbf{CLIP score}~\cite{clip} measures agreement between generated frames and the evaluation text. \textbf{UnifiedReward-2 Style (UR2-S)}~\cite{ur2} assesses aesthetic appeal, while \textbf{UnifiedReward-2 Alignment (UR2-A)}~\cite{ur2} measures agreement with the described content.

\textit{Visual and temporal coherence.} \textbf{CSD}~\cite{csd} measures style consistency between subsequent output frames and the first frame. \textbf{VBench Motion Smoothness (Smooth)}~\cite{vbench} evaluates motion smoothness on continuous output videos, while \textbf{UnifiedReward-2 Coherence (UR2-C)}~\cite{ur2} assesses within-frame visual coherence and distortions. For a fair comparison, outputs from all methods are resized to 480p and resampled to 8 fps for evaluation. Details of baseline implementations and metric computation are provided in the supplementary material.

\subsection{Comparison with Existing Methods}

\paragraph{Quantitative results.}
Table~\ref{tab:main} reports results across three categories. \textit{Content preservation.} \method and Diffutoon rank first and second in Line-F1 and UR-E, consistent with their explicit structure guidance and colorization design. \method also achieves the lowest Chroma error of 4.86, improving color preservation over Diffutoon's 9.30.

\textit{Style and text alignment.} \method leads all three metrics, supporting effective anime stylization with content alignment and aesthetic quality. AnyV2V and MiniMax-H3-Ref2VA obtain the lowest CLIP and UR2-A scores, consistent with their limited transformation of the source's photographic appearance.

\textit{Visual and temporal coherence.} \method achieves the highest CSD and UR2-C scores and ties Seedance2.5 in motion smoothness at the reported precision. These results support coherent rendering within and across frames, consistent with the video backbone's temporal prior and framewise source guidance.

\paragraph{Qualitative results.}
Figures~\ref{fig:teaser} and~\ref{fig:comparison} show that \method produces convincing anime renderings while preserving source outlines and colors. Seedance2.5 does not consistently stylize the entire scene, sometimes transforming only foreground objects or retaining a non-anime appearance throughout. EchoStyle, PickStyle, and Kiwi-Edit generate anime-style outputs but introduce larger changes to the source content and lose fine details. Diffutoon preserves source outlines well, but its image-based generation backbone provides weaker color consistency across frames, an observation consistent with its strong Line-F1 and lower CSD scores in Table~\ref{tab:main}. AnyV2V and MiniMax-H3-Ref2VA largely retain the original photographic appearance, as shown in Fig.~\ref{fig:comparison}.

\subsection{Ablation Studies}
\label{sec:ablation}
\paragraph{Structure guidance and colorization.}
Both ablation studies use Line-F1 for outline preservation, Chroma for color preservation, CLIP for text alignment, and CSD for consistency. Table~\ref{tab:ablation} and Fig.~\ref{fig:ablation} examine the two modules. Structure guidance primarily improves Line-F1, while colorization reduces Chroma error. Their combination at $(0.5,0.5)$ achieves the best outline preservation, color preservation, alignment, and consistency. We find that scale combinations near $(0.5,0.5)$ yield similar results, with further analysis provided in the supplementary material. Figure~\ref{fig:ablation} shows that increasing the structure scale preserves more details, whereas increasing the colorization scale produces colors closer to the source. Without Gaussian filtering during training and inference, the RGB-conditioned variant still retains source colors reasonably well, with a Chroma error of 9.65, but its CLIP score drops from 30.5 to 21.8. Its outputs retain a photographic appearance rather than the desired anime style, supporting the use of low-frequency color conditions.

\begin{table}[t]
\centering
\caption{Control and training ablations. The RGB variant is retrained with the same budget.}
\label{tab:ablation}
\scriptsize\setlength{\tabcolsep}{2pt}
\begin{tabularx}{\linewidth}{@{}l*{4}{Y}@{}}
\toprule
Setting $(\lambda_s,\lambda_c)$ & \mbox{Line-F1 $\uparrow$} & \mbox{Chroma $\downarrow$} & \mbox{CSD $\uparrow$} & \mbox{CLIP score $\uparrow$}\\
\midrule
$(0,0)$ & 0.29 & 28.87 & \underline{0.91} & 27.90\\
$(0.5,0)$ & 0.41 & 27.73 & \textbf{0.92} & 28.20\\
$(0,0.5)$ & 0.34 & 17.11 & \textbf{0.92} & 29.10\\
$(1.0,0)$ & 0.55 & 25.27 & \textbf{0.92} & 29.40\\
$(0,1.0)$ & 0.41 & 10.81 & 0.90 & 28.90\\
$(0.5,1.0)$ & 0.56 & 10.25 & 0.90 & 28.20\\
$(1.0,0.5)$ & \underline{0.60} & 14.40 & 0.90 & \underline{29.70}\\
RGB $(0.5,0.5)$ & 0.57 & \underline{9.65} & 0.89 & 21.80\\
\method $(0.5,0.5)$ & \textbf{0.64} & \textbf{4.86} & \textbf{0.92} & \textbf{30.50}\\
\bottomrule
\end{tabularx}
\end{table}

\paragraph{Consistency across video segments.}
We evaluate overlap length on the 21 videos in our 70-video evaluation set that exceed 328 frames after resampling to 24 fps. Table~\ref{tab:overlap_ablation} favors $O=1$. Larger overlaps weaken outline preservation, color preservation, and consistency, with independent generation performing worst in all three. Figure~\ref{fig:continuation} compares latent-context continuation with overlap lengths $O=1,22,56,90$ against independent-window generation. We align the first new source timestamp and fix the prompt, control strengths, resolution, and random seed across variants. Independent generation introduces abrupt appearance changes at the boundary. Longer contexts make the following window appear increasingly washed out with lower contrast, whereas a one-frame context better preserves the transition's appearance.

\begin{table}[t]
\centering
\caption{Overlap-length ablation on the 21 evaluation videos exceeding 328 frames at 24 fps. Bold and underlining mark the best and second-best estimates.}
\label{tab:overlap_ablation}
\scriptsize\setlength{\tabcolsep}{1.5pt}
\begin{tabularx}{\linewidth}{@{}l*{4}{Y}@{}}
\toprule
Setting & \mbox{Line-F1 $\uparrow$} & \mbox{Chroma $\downarrow$} & \mbox{CSD $\uparrow$} & \mbox{CLIP $\uparrow$}\\
\midrule
$O=1$ & \textbf{0.689} & \textbf{9.01} & \textbf{0.879} & \textbf{28.23}\\
$O=22$ & \underline{0.687} & \underline{9.03} & \underline{0.878} & 27.94\\
$O=56$ & 0.686 & 9.04 & 0.877 & 28.00\\
$O=90$ & 0.683 & 9.15 & 0.874 & 27.81\\
Independent & 0.676 & 9.40 & 0.860 & \underline{28.17}\\
\bottomrule
\end{tabularx}
\end{table}

\subsection{Human Evaluation}
\label{sec:human}
We randomly select 10 videos from the evaluation set and recruit 30 participants, yielding 300 pairwise comparisons in total. For each video, each participant is shown the output of \method alongside that of a randomly selected baseline, without revealing the method identities, and rates both results on a scale from 1 to 5. As shown in Table~\ref{tab:human_main}, \method outperforms all baselines in human preference and achieves the highest mean rating and the lowest standard deviation.

\begin{table}[t]
\centering
\caption{Human evaluation. Preference columns compare \method with the method in each row. Ties contribute equally to both preference columns. Mean and Std summarize ratings for that row's method, using the sample standard deviation. Pairs denotes the number of comparisons with \method. The \method row has no self-comparison preference or count.}
\label{tab:human_main}
\scriptsize\setlength{\tabcolsep}{2pt}
\begin{tabularx}{\linewidth}{@{}l*{5}{Y}@{}}
\toprule
\multirow{2}{*}[-1.5pt]{Method} & \multicolumn{2}{c}{Preference} & \multirow{2}{*}[-1.5pt]{Mean} & \multirow{2}{*}[-1.5pt]{Std} & \multirow{2}{*}[-1.5pt]{Pairs}\\
\cmidrule(lr){2-3}
& \method & Other & & & \\
\midrule
Diffutoon & 93.18\% & 6.82\% & 2.82 & 0.99 & 44\\
EchoStyle & 88.10\% & 11.90\% & 3.02 & 1.07 & 42\\
PickStyle & 94.74\% & 5.26\% & 2.71 & 0.96 & 38\\
Kiwi-Edit & 88.37\% & 11.63\% & 2.74 & \underline{0.95} & 43\\
AnyV2V & 97.62\% & 2.38\% & 1.79 & 1.00 & 42\\
MiniMax-H3-Ref2VA & 90.22\% & 9.78\% & 2.22 & 1.49 & 46\\
Seedance2.5 & 80.00\% & 20.00\% & \underline{3.07} & 1.03 & 45\\
\midrule
\method & - & - & \textbf{4.47} & \textbf{0.73} & -\\
\bottomrule
\end{tabularx}
\end{table}

\section{Conclusion}
We presented \method, a video diffusion framework for toon shading that combines structure guidance from outlines with colorization from low-frequency appearance. Both modules are trained independently through animation reconstruction, enabling anime rendering without cross-domain training pairs or external style references. Across 70 photorealistic videos, \method achieves leading results in content preservation, style and text alignment, and visual and temporal coherence against specialized methods and general-purpose baselines. Human evaluation further favors \method over every baseline. Ablation studies confirm that complementary structure and low-frequency color guidance preserve source content while enabling anime rendering.

\section*{LLM Usage Statement}
OpenAI's ChatGPT and Codex were utilized to support language polishing and grammatical improvements across the manuscript, along with LaTeX formatting and the refinement of figure layout and labeling. These tools were not employed to produce experimental data, numerical results, or references. The authors have verified and approved all scientific statements, equations, analyses, and final content, and bear full responsibility for the submission.

{\small\bibliographystyle{ieeenat_fullname}\bibliography{main}}

\clearpage
\appendix
\twocolumn[\begin{center}
{\Large\bfseries Diffutoon-2: Toon Shading with Video Diffusion Models\\[6pt]Supplementary Material}
\end{center}\vspace{12pt}]
\section{Training and Inference Details}
\label{app:implementation}
\paragraph{Training.}
We train the structure and colorization modules separately on the same private dataset of 15,605 animation clips. Each clip is paired with its extracted outlines for structure training and its Gaussian-filtered version for colorization training. The original animation is the reconstruction target in both cases. Each run updates one module with flow matching while keeping the MiniMax-H3 backbone and encoders frozen; the other module is absent. Both modules use nine control blocks corresponding to backbone layers $0,5,10,15,20,25,30,35,40$.

We use 39-frame clips, a batch size of 8, and a learning rate of $10^{-5}$. Training preserves the source aspect ratio and uses the largest supported resolution below one million pixels. Each module is trained for 5,000 steps on eight PPU-ZW810s with 96 GB of memory each, taking approximately 36 hours.

\begin{table*}[t]
\centering\small
\caption{Training settings shared by the two modules. Each module is optimized in a separate run.}
\label{tab:training}
\begin{tabularx}{\linewidth}{@{}l*{2}{Y}@{}}
\toprule
Setting & Structure module & Colorization module\\\midrule
Training data & \multicolumn{2}{c}{15,605 animation clips}\\
Condition & Extracted outlines & Gaussian-filtered video\\
Target & Original animation & Original animation\\
Training text & $p_s$ & $p_c$\\
Objective & Flow matching & Flow matching\\
Trainable parameters & 6.1B & 6.1B\\
Backbone and encoders & Frozen & Frozen\\
Steps & 5,000 & 5,000\\
Batch size & 8 & 8\\
Learning rate & $10^{-5}$ & $10^{-5}$\\
Clip length & 39 frames & 39 frames\\
Resolution & \multicolumn{2}{c}{Original aspect ratio, at most one million pixels}\\
Hardware & \multicolumn{2}{c}{Eight PPU-ZW810s, each with 96 GB memory}\\
Training time & Approximately 36 hours & Approximately 36 hours\\
\bottomrule
\end{tabularx}
\end{table*}

\paragraph{Preprocessing and sampling.}
Input videos are resampled to 24 fps. Output dimensions preserve the input aspect ratio under a budget of 1,032,192 pixels and are aligned to multiples of 32. The structure condition is extracted with Informative Drawings~\cite{informative_drawings}, using a short side of 512 pixels before resizing the outlines back to the source dimensions. The color condition uses a spatial Gaussian filter with $\sigma=20$ in source pixels. Both control strengths are 0.5. We use the four-step DMD LoRA supplied by LightX2V~\cite{lightx2v}.

\paragraph{Window continuation.}
We use an overlap of $O=1$ frame. The final window is rounded up to the smallest supported length $17n+5$, where $n\geq1$, and excess output frames are discarded. MiniMax-H3 encodes each 17-frame block into five temporal latent positions, with one for the first frame and one for each subsequent group of four frames. Our one-frame overlap therefore corresponds to a single temporal latent position.

\section{Textual Conditions and Caption Generation}
\label{app:text}
We use Qwen3.8-27B~\cite{qwen38} to describe video content and combine the descriptions with task instructions. Training uses the respective text conditions $p_s$ and $p_c$ for structure and colorization. At inference, a fixed style instruction is combined with the description of the current video window. Prompts~\ref{prompt:structure} and~\ref{prompt:colorization} specify the two training tasks, and Prompt~\ref{prompt:rendering} gives the inference text wrapper.

\paragraph{Training descriptions.}
The training metadata uses a fixed wrapper for each module, with only \texttt{\{clip\_caption\}} changing between clips. Prompt~\ref{prompt:structure} treats the outlines as the complete specification of scene content and motion. Prompt~\ref{prompt:colorization} asks for clean anime appearance while retaining the blurred video's composition and temporal structure.

\paragraph{Global and local inference captions.}
For the global description, we uniformly sample up to 12 frames across the source video. For each local description, we sample up to 12 frames from the corresponding window, excluding the retained overlap in later windows. Sampling follows the output timeline at 24 fps, with output frame $k$ mapped to source frame $\operatorname{round}(k f_s/24)$ for source rate $f_s$. Sampling stops at the source endpoint, so padded generation frames are never captioned. Images with a short side above 560 pixels are downscaled while preserving their aspect ratio.

Prompt~\ref{prompt:system} supplies the common system instruction. Prompt~\ref{prompt:global} establishes the subjects, setting, and major changes. The local instruction in Prompt~\ref{prompt:local} receives the current frames and is extended with the position and reference blocks in Prompts~\ref{prompt:position} and~\ref{prompt:reference}. The latter provides a common naming reference; local appearance and actions must still be supported by the current frames. Qwen receives a shared system instruction followed by the images in temporal order and the user prompt. We use bfloat16, medium reasoning effort, and a maximum of 2,048 new tokens. The global and local prompts request at most 150 words. The parsed local description fills Prompt~\ref{prompt:rendering}; the global description is used only to guide local captioning.

\clearpage
\onecolumn
\noindent The numbered prompt templates below preserve the training wrappers and inference instructions. Braced fields are substituted with the relevant descriptions or video metadata. Punctuation is reproduced verbatim.

\promptfile{prompt:structure}{Structure-training prompt $p_s$. The field \texttt{\{clip\_caption\}} is replaced with the description of animation clip $A$.}{prompts/training_structure.txt}
\promptfile{prompt:colorization}{Colorization-training prompt $p_c$, using the same per-clip description field.}{prompts/training_colorization.txt}
\promptfile{prompt:system}{Shared system instruction for Qwen3.8-27B inference captioning.}{prompts/system_prompt_display.txt}
\promptfile{prompt:global}{Global-caption user instruction. The fields \texttt{\{n\}} and \texttt{\{dur\}} specify the sampled-frame count and video duration.}{prompts/global_prompt_display.txt}
\promptfile{prompt:local}{Local-caption user instruction, populated with the current segment's sampled-frame count and duration.}{prompts/user_prompt_display.txt}
\promptfile{prompt:position}{Position information appended to the local-caption instruction.}{prompts/position_block_display.txt}
\promptfile{prompt:reference}{Global-description reference appended to the local-caption instruction.}{prompts/local_block_display.txt}
\promptfile{prompt:rendering}{Video-generation prompt. The field \texttt{\{local\_caption\}} is replaced with the parsed description of the current window.}{prompts/inference_wrapper.txt}

\clearpage
\twocolumn

\section{Baseline Implementation Details}
\label{app:baselines}
Before computing any metric, we resize videos from every method to 480p and sample frames at 8 fps to ensure a fair comparison. Source-dependent metrics use source and generated frames aligned at the same timestamps and spatial coordinates. Whenever a baseline uses a VLM for caption generation, we use Qwen3.8-27B~\cite{qwen38}. All other components and settings follow the original methods. Evaluation text and metric-model settings are shared across methods. MiniMax-H3-Ref2VA receives the outline and blurred color videos as dual video references and uses four denoising steps. Its spatial budget and frame rate match ours. Diffutoon uses ten denoising steps in its main toon-shading pipeline. Table~\ref{tab:baseline_settings} summarizes model sizes and generation settings.

\begin{table}[t]
\centering
\caption{Model sizes and generation settings before evaluation resizing and frame sampling. Parameters denote model size rather than task-specific trainable parameters. Source FPS follows the input video; N/A denotes undisclosed values.}
\label{tab:baseline_settings}
\scriptsize\setlength{\tabcolsep}{2pt}
\begin{tabularx}{\linewidth}{@{}lcYcc@{}}
\toprule
Method & Parameters & Resolution & FPS & Steps\\
\midrule
Diffutoon & 2.243B & $896\times512$ & Source & 10\\
EchoStyle & 35.009B & $832\times480$ & 16 & 40\\
PickStyle & 23.177B & $480\times832$ & 16 & 20\\
Kiwi-Edit & 9.477B & $1280\times704$ & Source & 50\\
AnyV2V & 3.555B & $1280\times704$ & 8 & 50\\
MiniMax-H3-Ref2VA & 69.927B & $1344\times768$ & 24 & 4\\
Seedance2.5 & N/A & $1280\times720$ & 24 & N/A\\
\midrule
\method & 82.128B & $1344\times768$ & 24 & 4\\
\bottomrule
\end{tabularx}
\end{table}

\section{Evaluation Metrics}
\label{app:metrics}
Let $x_i$ and $y_i$ denote the aligned source and generated frames, respectively, and let $N$ be the number of evaluated frames in a video.

\paragraph{Line-art Boundary F1-score (Line-F1).}
We extract source and output line drawings with the same Informative Drawings detector~\cite{informative_drawings}. Each line-art video is binarized using its own Otsu threshold, estimated from sampled frames and held fixed across the sequence. For frame $i$, we compute boundary precision and recall with a spatial tolerance,
\begin{equation}
\begin{aligned}
 P_i&=\frac{|\{q\in\mathcal{B}_i^y:d(q,\mathcal{B}_i^x)\leq\delta\}|}
 {|\mathcal{B}_i^y|},\\
 R_i&=\frac{|\{q\in\mathcal{B}_i^x:d(q,\mathcal{B}_i^y)\leq\delta\}|}
 {|\mathcal{B}_i^x|},\\
 \mathrm{Line\text{-}F1}&=\frac{1}{N}\sum_{i=1}^{N}\frac{2P_iR_i}{P_i+R_i}.
\end{aligned}
\end{equation}
Here $\mathcal{B}_i^x$ and $\mathcal{B}_i^y$ are the source and generated boundary-pixel sets, $q$ is a pixel location, $|\cdot|$ counts pixels, and $d(q,\mathcal{B})$ is the Euclidean distance to the nearest pixel in set $\mathcal{B}$. We set $\delta=1.5$ pixels, so a boundary pixel can match the same location or any of its eight immediate neighbors, including diagonal neighbors at distance $\sqrt{2}$. $P_i$ and $R_i$ are the resulting precision and recall. A frame receives zero when $P_i+R_i=0$ or only one boundary map is empty; it receives one when both maps are empty. Higher scores indicate better outline preservation.

\paragraph{Chroma.}
We adapt CIEDE2000 by smoothing the CIELAB chroma channels and removing the lightness difference. Define
\begin{equation}
\begin{aligned}
 \Psi(x)&=\big[50,\ G_{\sigma_c}(a(x)),\ G_{\sigma_c}(b(x))\big],\\
 \mathrm{Chroma}&=\frac{1}{N|\Omega|}
 \sum_{i=1}^{N}\sum_{q\in\Omega}
 \Delta E_{00}\big(\Psi(x_i)_q,\Psi(y_i)_q\big).
\end{aligned}
\end{equation}
Here $\Psi$ is the transformed Lab representation, $a(x)$ and $b(x)$ are its chroma channels, and $G_{\sigma_c}$ is a spatial Gaussian filter with standard deviation $\sigma_c$. Both lightness channels are set to the same constant, chosen here as 50. Any common value produces the same score because the CIEDE2000 lightness-difference term vanishes; 50 serves only as a placeholder for this channel. The set $\Omega$ contains the evaluated pixel positions, $|\Omega|$ is their count, and $\Delta E_{00}$ denotes CIEDE2000 between two transformed pixel colors. The subscript $q$ selects a pixel. This score preserves sensitivity to spatial color changes while allowing cel shading to change luminance. Lower values are better. The evaluation smoothing scale $\sigma_c$ is separate from the blur used to construct the model's color condition.

\paragraph{UnifiedReward-Edit (UR-E).}
We use the pointwise editing evaluator with the source frame, generated frame, and editing instruction. It returns two scores from 0 to 25 for editing success and source preservation against unnecessary changes. UR-E is their arithmetic mean, averaged over evaluated frames. The preservation component assigns higher scores to better preservation.

\paragraph{CLIP score.}
We use CLIP-ViT-H-14 with the \texttt{laion2B-s32B-b79K} weights~\cite{clip}. The implemented score is
\begin{equation}
 \mathrm{CLIP}=\frac{1}{N}\sum_{i=1}^{N}
 e^{\ell}\frac{f_I(y_i)^\top f_T(p_i)}
 {\|f_I(y_i)\|_2\,\|f_T(p_i)\|_2},
\end{equation}
where $f_I$ and $f_T$ are the image and text encoders, $p_i$ is the evaluation text associated with frame $i$, and $\ell$ is the checkpoint's learned logit-scale parameter. The transpose and $\|\cdot\|_2$ denote the inner product and Euclidean norm. We retain the checkpoint's scale $e^{\ell}$, so the reported score is a scaled similarity rather than a probability. Text is tokenized with a maximum length of 77 tokens.

\paragraph{UnifiedReward-2 Style (UR2-S), Alignment (UR2-A), and Coherence (UR2-C).}
UnifiedReward-2~\cite{ur2} evaluates each generated frame together with its text description in a single VLM call. The three outputs are scored separately from 1 to 5. UR2-S measures aesthetic appeal, UR2-A measures content agreement with the text, and UR2-C measures within-frame visual coherence, including object distortions and visual glitches. We report each axis separately without combining them into a total. UR2-C evaluates individual frames, not temporal consistency.

\paragraph{CSD.}
We use the official CSD descriptor~\cite{csd} to compare each evaluated output frame with the first frame,
\begin{equation}
 \mathrm{CSD}=\frac{1}{N-1}\sum_{i=2}^{N}
 \frac{f_S(y_1)^\top f_S(y_i)}
 {\|f_S(y_1)\|_2\,\|f_S(y_i)\|_2},
\end{equation}
where $f_S$ is the CSD style encoder, $y_1$ is the initial generated frame, and $N>1$. Each term is the cosine similarity of two style descriptors. The first frame's self-similarity is excluded. This measures consistency of the rendered style within a video and requires no external style reference.

\paragraph{VBench Motion Smoothness (Smooth).}
We use the official VBench frame-interpolation evaluator~\cite{vbench}. It retains alternate frames, reconstructs the intermediate frames with its interpolation model, and compares them with the actual held-out frames,
\begin{equation}
 \mathrm{Smooth}=1-\frac{1}{255M}
 \sum_{m=1}^{M}\operatorname{MAE}(y_{i_m},\widetilde y_{i_m}),
\end{equation}
where $M$ is the number of reconstructed intermediate frames, $i_m$ is the corresponding frame index, and $\widetilde y_{i_m}$ is its interpolated prediction. $\operatorname{MAE}$ averages the absolute difference over pixels and RGB channels on the 0 to 255 intensity scale. Higher scores indicate greater agreement with the interpolation model's motion prediction. The evaluator receives the complete ordered 8 fps sequence after the common preprocessing.

\section{Consistency of Global and Local Captions}
\label{app:caption_consistency}
Figures~\ref{fig:caption_case_66} and~\ref{fig:caption_case_91} compare descriptions of two adjacent windows with and without a global reference. Our procedure reuses names for recurring subjects and settings, while allowing each window to describe its own events. In these examples, global guidance retains ``heathland'' and ``tall tapering skyscraper'' across windows. Independently generated captions vary the scene description or introduce a specific landmark name in the second window. These excerpts illustrate greater naming consistency; differences in wording alone do not establish incorrect scene content. We show the full global descriptions and the relevant opening sentences of the local captions.

\begin{figure*}[p]
\centering
\begin{minipage}{\textwidth}
\begin{tcolorbox}[promptbox,title={Sheep in scrubland \quad PixaBay100 video 66},colbacktitle=black!8,coltitle=black,fonttitle=\bfseries\small]
\small\textbf{Global caption}\par In an open heathland of low dry scrub, with a distant line of green shrubs and dark trees under a pale sky, two white sheep rest in the mid-distance. The view then tightens to the edge of the vegetation, where a small white lamb with pointed ears moves among the low branches, its mother\textemdash{}a larger sheep with shaggy white wool\textemdash{}standing nearby. Over the sequence the lamb shifts position and the adult sheep walks slowly rightward through the scrub, the lamb staying close at its side. Toward the end the adult moves farther right and becomes partly hidden by foreground branches, while the lamb drifts into the shrubs at the right edge of the frame.
\end{tcolorbox}
\small\setlength{\tabcolsep}{7pt}
\begin{tabularx}{\textwidth}{@{}XX@{}}
\toprule
\multicolumn{2}{c}{\textbf{Global + local (ours)}}\\
\textbf{First window} & \textbf{Second window}\\
The segment opens on a wide \captionmatch{heathland}: low dry scrub in the foreground, a band of pale grass and green shrubs, and a dark tree line under a pale sky, with two white sheep resting in the mid-distance. & In an open \captionmatch{heathland} of low dry scrub under a pale sky with a distant green tree line, a large shaggy white sheep stands in the mid-distance, head turned slightly toward the camera, with a small white lamb with pointed ears at the right edge among the branches.\\
\midrule
\multicolumn{2}{c}{\textbf{Local caption only}}\\
\textbf{First window} & \textbf{Second window}\\
The segment opens on a wide view of a \captiondrift{moorland} bordered by a dark pine forest, where two white sheep rest on the ground. & A sheep with long, shaggy cream-colored wool stands in an \captiondrift{open field of dry, low brush} under a pale dawn or dusk sky, a low tree line stretching across the background.\\
\bottomrule\end{tabularx}
\end{minipage}
\caption{The same setting is described consistently as heathland with a global reference, while independent local captions use different scene descriptions. The global caption is shown in full; window captions are opening-sentence excerpts. Green bold text marks repeated wording, and red bold text marks differing wording.}
\label{fig:caption_case_66}
\end{figure*}

\begin{figure*}[p]
\centering
\begin{minipage}{\textwidth}
\begin{tcolorbox}[promptbox,title={City skyline \quad PixaBay100 video 91},colbacktitle=black!8,coltitle=black,fonttitle=\bfseries\small]
\small\textbf{Global caption}\par A wide city skyline, dominated by a tall tapering skyscraper and a cluster of high-rises, rises above a flat expanse of low-rise buildings under a hazy sky. The scene is a sunset timelapse: the sky begins as a bright, glowing orange-gold with soft horizontal cloud bands, then the clouds gradually thicken and the light deepens toward a richer amber. In the final stretch the upper sky darkens into muted grey-purple, the horizon fades, and small artificial lights begin to appear as scattered white and warm dots across the lower city. The skyline silhouette itself remains fixed throughout, serving as a constant against the shifting sky.
\end{tcolorbox}
\small\setlength{\tabcolsep}{7pt}
\begin{tabularx}{\textwidth}{@{}XX@{}}
\toprule
\multicolumn{2}{c}{\textbf{Global + local (ours)}}\\
\textbf{First window} & \textbf{Second window}\\
A wide city skyline, anchored by a \captionmatch{tall tapering skyscraper} and a dense cluster of high-rises, rises above a flat field of low-rise buildings. & A wide city skyline, led by a \captionmatch{tall tapering skyscraper} and a dense cluster of high-rises, stands fixed above a flat expanse of low-rise rooftops.\\
\midrule
\multicolumn{2}{c}{\textbf{Local caption only}}\\
\textbf{First window} & \textbf{Second window}\\
A wide view of a modern city skyline, dominated by a tall \captiondrift{needle-like tower} at the center, stretches across the middle of the frame above a dense foreground of low buildings. & A wide view of the Dubai skyline at dusk, with the \captiondrift{Burj Khalifa} rising as the tallest spire among a cluster of dark, silhouetted skyscrapers.\\
\bottomrule\end{tabularx}
\end{minipage}
\caption{The global reference maintains the same descriptive name for the main building, while local-only captions shift from a generic tower description to a named landmark. The global caption is shown in full; window captions are opening-sentence excerpts. Green bold text marks repeated wording, and red bold text marks differing wording.}
\label{fig:caption_case_91}
\end{figure*}

\clearpage

\section{Sensitivity to Control Scales}
\label{app:scales}
\begin{figure*}[p]
\centering
\includegraphics[width=\textwidth,height=0.9\textheight,keepaspectratio]{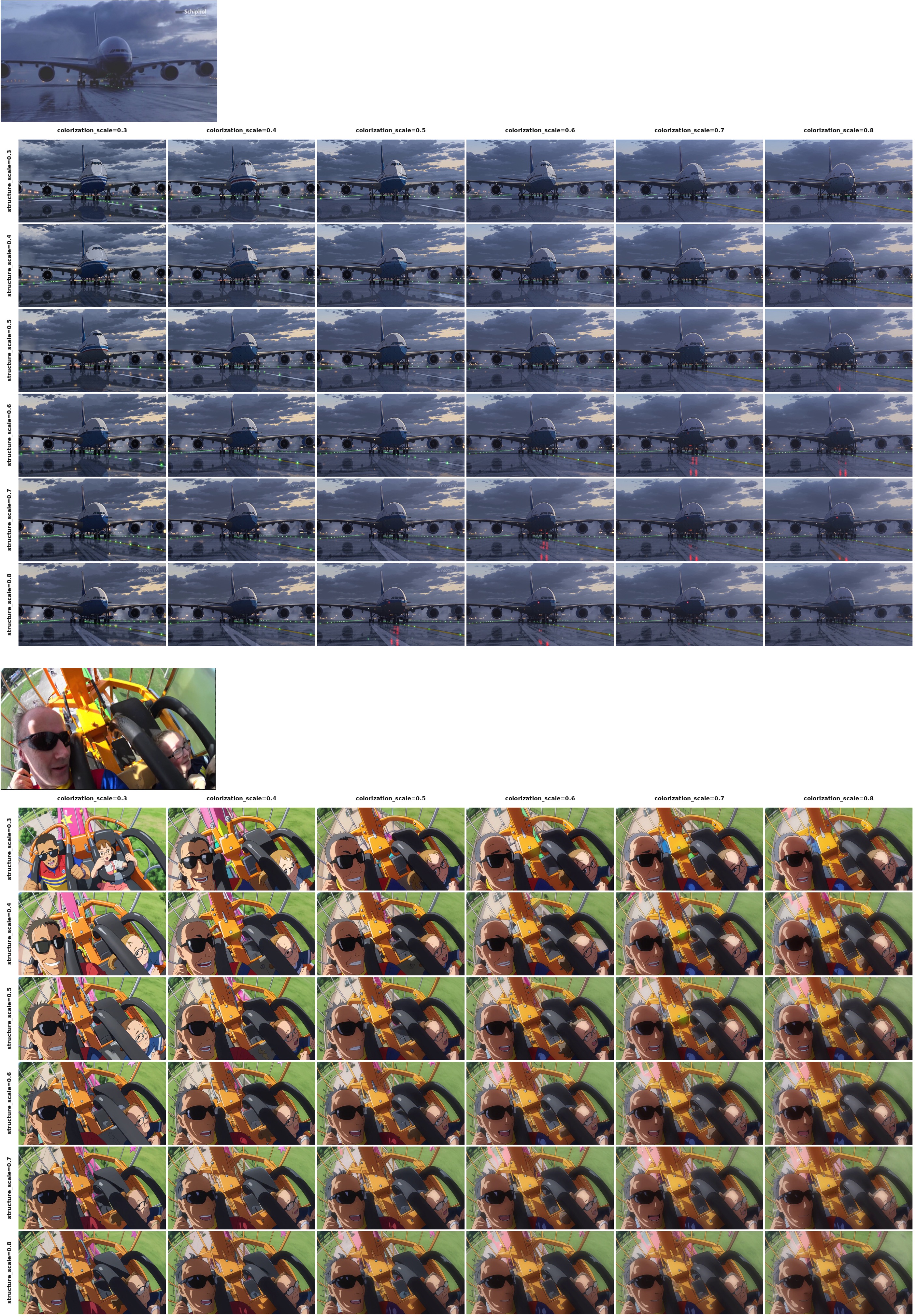}
\caption{Sensitivity to structure and colorization scales. Each example shows its source frame above a $6\times6$ grid. Rows and columns vary $\lambda_s$ and $\lambda_c$, respectively, from 0.3 to 0.8. Settings near $(0.5,0.5)$ yield similar overall results.}
\label{fig:scale_sweep}
\end{figure*}

Figure~\ref{fig:scale_sweep} shows combinations of structure and colorization scales from 0.3 to 0.8 in increments of 0.1. Rows vary the structure scale $\lambda_s$, and columns vary the colorization scale $\lambda_c$. Combinations near $(0.5,0.5)$ produce similar overall appearance, supporting our default setting without requiring precise tuning. Increasing structure guidance generally retains more source details, while stronger colorization brings the palette closer to the input.

\clearpage
\section{Sensitivity to Gaussian Blur Strength}
\label{app:sigma}
\begin{figure*}[t]
\centering
\includegraphics[width=\textwidth]{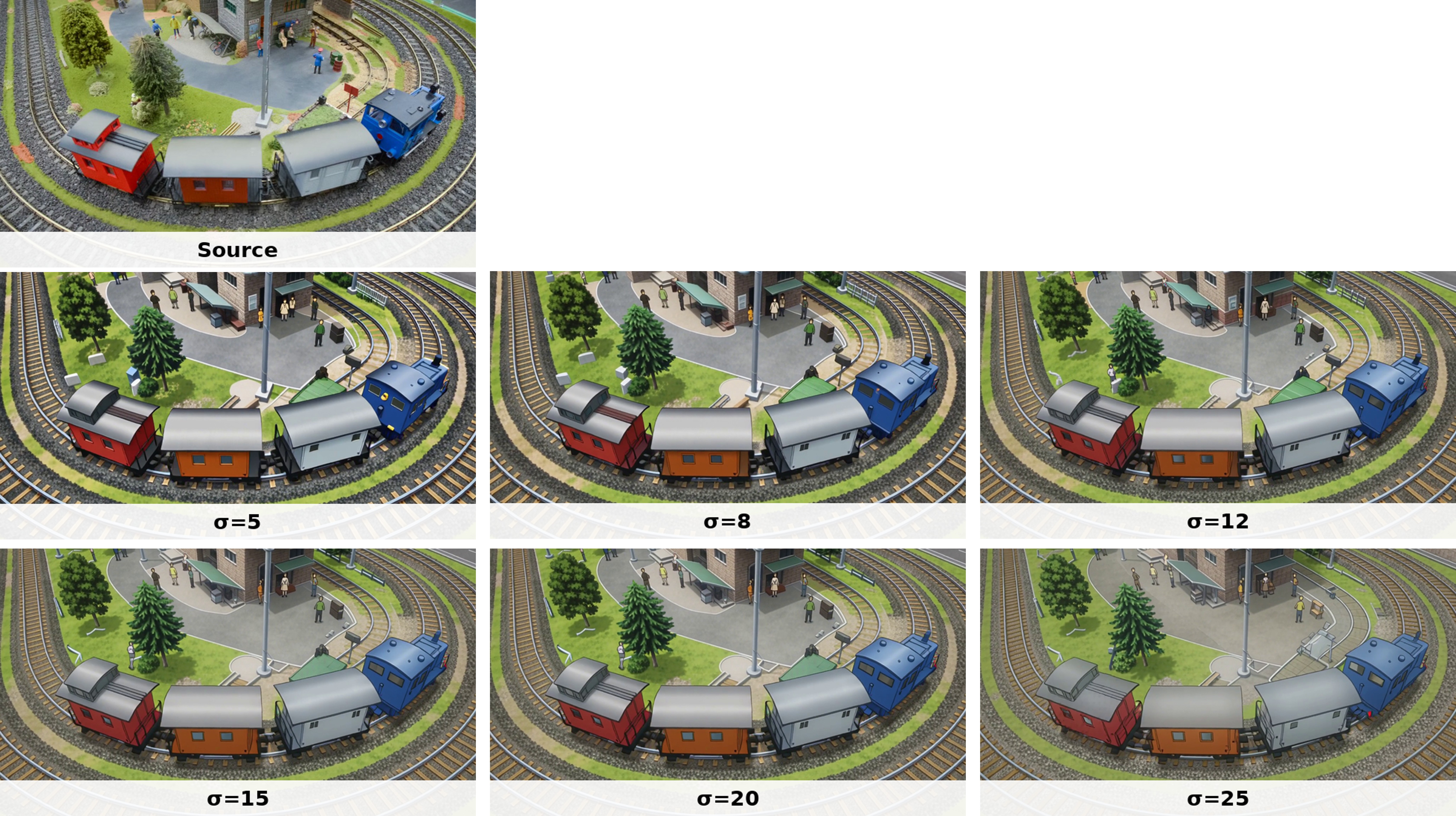}
\caption{Effect of Gaussian blur strength at inference. The source frame is followed by results with $\sigma=5,8,12,15,20,25$. Small values yield weaker cel shading, while excessive blur softens the rendered details.}
\label{fig:sigma_sweep}
\end{figure*}

We vary the Gaussian blur standard deviation at inference over $\sigma\in\{5,8,12,15,20,25\}$. Figure~\ref{fig:sigma_sweep} shows that small values lead to insufficient stylization, with smooth color gradients rather than the distinct, flat color regions characteristic of cel shading. Excessively large values produce a softer, slightly blurred appearance. Table~\ref{tab:sigma_ablation} compares the six settings on all 70 evaluation videos using all nine metrics from the main evaluation.

\begin{table*}[t]
\centering
\caption{Gaussian blur strength ablation on 70 evaluation videos.}
\label{tab:sigma_ablation}
\scriptsize\setlength{\tabcolsep}{1.5pt}
\begin{tabularx}{\linewidth}{@{}l*{9}{Y}@{}}
\toprule
\multirow{2}{*}{$\sigma$}
& \multicolumn{3}{c}{\scriptsize Content Preservation}
& \multicolumn{3}{c}{\scriptsize Style and Text Alignment}
& \multicolumn{3}{c}{\scriptsize Visual and Temporal Coherence}\\
\cmidrule(lr){2-4}\cmidrule(lr){5-7}\cmidrule(lr){8-10}
& \mbox{Line-F1 $\uparrow$} & \mbox{Chroma $\downarrow$} & \mbox{UR-E $\uparrow$}
& \mbox{CLIP $\uparrow$} & \mbox{UR2-S $\uparrow$} & \mbox{UR2-A $\uparrow$}
& \mbox{CSD $\uparrow$} & \mbox{Smooth. $\uparrow$} & \mbox{UR2-C $\uparrow$}\\
\midrule
5 & 0.65 & 4.74 & 22.88 & 29.45 & 3.38 & 3.40 & 0.91 & 0.98 & 3.84\\
8 & 0.64 & 4.75 & 22.97 & 29.95 & 3.38 & 3.41 & 0.91 & 0.98 & 3.85\\
12 & 0.64 & 4.83 & 22.95 & 30.15 & 3.38 & 3.41 & 0.91 & 0.98 & 3.84\\
15 & 0.64 & 4.87 & 22.97 & 30.30 & 3.38 & 3.39 & 0.91 & 0.98 & 3.86\\
20 & 0.64 & 4.86 & 23.00 & 30.50 & 3.38 & 3.39 & 0.92 & 0.99 & 3.87\\
25 & 0.64 & 4.99 & 23.01 & 30.96 & 3.36 & 3.44 & 0.92 & 0.99 & 3.86\\
\bottomrule
\end{tabularx}
\end{table*}

\clearpage
\onecolumn
\section{Seventy-Video Evaluation Manifest}
Table~\ref{tab:video_manifest} lists the 30 PixaBay100 and 40 DAVIS2017 source videos used for evaluation. Resolution, frame rate, duration, and frame count describe the source files before generation and evaluation resampling. DAVIS videos are selected from the combined training and validation sets.
\small
\begin{longtable}{@{}rllrrrr@{}}
\caption{Source-video metadata for the evaluation set.}\label{tab:video_manifest}\\
\toprule
ID & Dataset & Source ID / name & Resolution & FPS & Duration (s) & Frames\\\midrule
\endfirsthead
\toprule
ID & Dataset & Source ID / name & Resolution & FPS & Duration (s) & Frames\\\midrule
\endhead
1 & PixaBay100 & 8 & $1920\times1080$ & 25 & 16.00 & 400\\
2 & PixaBay100 & 10 & $1920\times1080$ & 29.97 & 20.79 & 623\\
3 & PixaBay100 & 12 & $1920\times1080$ & 29.97 & 25.12 & 753\\
4 & PixaBay100 & 13 & $1920\times1080$ & 30 & 25.10 & 753\\
5 & PixaBay100 & 14 & $1920\times1080$ & 29.97 & 25.36 & 760\\
6 & PixaBay100 & 22 & $1920\times1080$ & 25 & 8.56 & 214\\
7 & PixaBay100 & 24 & $1920\times1080$ & 30 & 11.03 & 331\\
8 & PixaBay100 & 30 & $1920\times1080$ & 23.976 & 16.18 & 388\\
9 & PixaBay100 & 32 & $1920\times1080$ & 29.97 & 15.38 & 461\\
10 & PixaBay100 & 33 & $1920\times1080$ & 25 & 30.52 & 763\\
11 & PixaBay100 & 34 & $1920\times1080$ & 29.97 & 25.06 & 751\\
12 & PixaBay100 & 35 & $1920\times1080$ & 29.97 & 9.81 & 294\\
13 & PixaBay100 & 36 & $1920\times1080$ & 100 & 45.53 & 4553\\
14 & PixaBay100 & 46 & $1920\times1080$ & 25 & 7.04 & 176\\
15 & PixaBay100 & 47 & $1920\times1080$ & 30 & 19.80 & 594\\
16 & PixaBay100 & 48 & $1920\times1080$ & 23.976 & 9.84 & 236\\
17 & PixaBay100 & 55 & $1920\times1080$ & 29.97 & 9.01 & 270\\
18 & PixaBay100 & 61 & $1920\times1080$ & 23.976 & 21.65 & 519\\
19 & PixaBay100 & 66 & $1920\times1080$ & 29.97 & 18.75 & 562\\
20 & PixaBay100 & 70 & $1920\times1080$ & 29.97 & 11.34 & 340\\
21 & PixaBay100 & 73 & $1920\times1080$ & 59.94 & 25.53 & 1530\\
22 & PixaBay100 & 78 & $1920\times1080$ & 23.976 & 17.98 & 431\\
23 & PixaBay100 & 80 & $1920\times1080$ & 29.97 & 12.88 & 386\\
24 & PixaBay100 & 85 & $1920\times1080$ & 23.976 & 19.81 & 475\\
25 & PixaBay100 & 91 & $1920\times1080$ & 25 & 15.96 & 399\\
26 & PixaBay100 & 92 & $2048\times1080$ & 23.976 & 18.56 & 445\\
27 & PixaBay100 & 98 & $1920\times1080$ & 23.976 & 20.23 & 485\\
28 & PixaBay100 & 100 & $1920\times1080$ & 24 & 18.75 & 450\\
29 & PixaBay100 & 103 & $1920\times1080$ & 29.97 & 8.01 & 240\\
30 & PixaBay100 & 105 & $1920\times1080$ & 25 & 20.68 & 517\\
31 & DAVIS2017 & bike-packing & $910\times480$ & 24 & 2.88 & 69\\
32 & DAVIS2017 & boat & $854\times480$ & 24 & 3.12 & 75\\
33 & DAVIS2017 & breakdance-flare & $854\times480$ & 24 & 2.96 & 71\\
34 & DAVIS2017 & bus & $854\times480$ & 24 & 3.33 & 80\\
35 & DAVIS2017 & car-roundabout & $854\times480$ & 24 & 3.12 & 75\\
36 & DAVIS2017 & car-shadow & $854\times480$ & 24 & 1.67 & 40\\
37 & DAVIS2017 & car-turn & $854\times480$ & 24 & 3.33 & 80\\
38 & DAVIS2017 & cat-girl & $912\times480$ & 24 & 3.71 & 89\\
39 & DAVIS2017 & cows & $854\times480$ & 24 & 4.33 & 104\\
40 & DAVIS2017 & disc-jockey & $1138\times480$ & 24 & 3.17 & 76\\
41 & DAVIS2017 & drift-straight & $854\times480$ & 24 & 2.08 & 50\\
42 & DAVIS2017 & gold-fish & $854\times480$ & 24 & 3.25 & 78\\
43 & DAVIS2017 & hike & $854\times480$ & 24 & 3.33 & 80\\
44 & DAVIS2017 & hockey & $854\times480$ & 24 & 3.12 & 75\\
45 & DAVIS2017 & horsejump-high & $854\times480$ & 24 & 2.08 & 50\\
46 & DAVIS2017 & kite-surf & $854\times480$ & 24 & 2.08 & 50\\
47 & DAVIS2017 & kite-walk & $854\times480$ & 24 & 3.33 & 80\\
48 & DAVIS2017 & koala & $854\times480$ & 24 & 4.17 & 100\\
49 & DAVIS2017 & lab-coat & $854\times480$ & 24 & 1.96 & 47\\
50 & DAVIS2017 & lady-running & $854\times480$ & 24 & 2.71 & 65\\
51 & DAVIS2017 & libby & $854\times480$ & 24 & 2.04 & 49\\
52 & DAVIS2017 & lucia & $854\times480$ & 24 & 2.92 & 70\\
53 & DAVIS2017 & mallard-fly & $854\times480$ & 24 & 2.92 & 70\\
54 & DAVIS2017 & mbike-trick & $854\times480$ & 24 & 3.29 & 79\\
55 & DAVIS2017 & motocross-bumps & $854\times480$ & 24 & 2.50 & 60\\
56 & DAVIS2017 & motorbike & $854\times480$ & 24 & 1.79 & 43\\
57 & DAVIS2017 & night-race & $854\times480$ & 24 & 1.92 & 46\\
58 & DAVIS2017 & paragliding-launch & $854\times480$ & 24 & 3.33 & 80\\
59 & DAVIS2017 & planes-water & $854\times480$ & 24 & 1.58 & 38\\
60 & DAVIS2017 & rallye & $854\times480$ & 24 & 2.08 & 50\\
61 & DAVIS2017 & rhino & $854\times480$ & 24 & 3.75 & 90\\
62 & DAVIS2017 & rollerblade & $854\times480$ & 24 & 1.46 & 35\\
63 & DAVIS2017 & scooter-gray & $854\times480$ & 24 & 3.12 & 75\\
64 & DAVIS2017 & sheep & $854\times480$ & 24 & 2.83 & 68\\
65 & DAVIS2017 & skate-park & $854\times480$ & 24 & 3.33 & 80\\
66 & DAVIS2017 & stroller & $854\times480$ & 24 & 3.79 & 91\\
67 & DAVIS2017 & swing & $854\times480$ & 24 & 2.50 & 60\\
68 & DAVIS2017 & train & $854\times480$ & 24 & 3.33 & 80\\
69 & DAVIS2017 & upside-down & $854\times480$ & 24 & 2.71 & 65\\
70 & DAVIS2017 & walking & $854\times480$ & 24 & 3.00 & 72\\
\bottomrule
\end{longtable}

\end{document}